\documentclass[journal]{IEEEtran}
\usepackage{cite}

\ifCLASSINFOpdf
   \usepackage[pdftex]{graphicx}
\else
\fi
\usepackage{amsmath}
\usepackage{url}

\usepackage{multirow}
\usepackage{mdwtab}
\usepackage{hhline}
\usepackage{epstopdf}

\begin{document}
%
\title{Objective Micro-Facial Movement Detection Using FACS-Based Regions and Baseline Evaluation}
%
%
%

\author{Adrian~K.~Davison,~\IEEEmembership{Member,~IEEE,}
        Cliff~Lansley,
        Choon~Ching~Ng,
        Kevin~Tan,
        Moi~Hoon~Yap,~\IEEEmembership{Member,~IEEE}
        \IEEEcompsocitemizethanks{\IEEEcompsocthanksitem {A. K. Davison, K. Tan and M.H. Yap are with the Informatics Research Centre, School of Computing, Mathematics and Digital Technology, Manchester Metropolitan University, Manchester, United Kingdom.
        		
        		Email: M.Yap@mmu.ac.uk
        		\IEEEcompsocthanksitem C. C. Ng is with the Panasonic R\&D Center, Singapore.
        		\IEEEcompsocthanksitem C. Lansley is with the Emotional Intelligence Academy, Walkden, United Kingdom.
        		}}%
}

\maketitle
\begin{abstract}
Micro-facial expressions are regarded as an important human behavioural event that can highlight emotional deception. Spotting these movements is difficult for humans and machines, however research into using computer vision to detect subtle facial expressions is growing in popularity. This paper proposes an individualised baseline micro-movement detection method using 3D Histogram of Oriented Gradients (3D HOG) temporal difference method. We define a face template consisting of 26 regions based on the Facial Action Coding System (FACS). We extract the temporal features of each region using 3D HOG. Then, we use Chi-square distance to find subtle facial motion in the local regions. Finally, an automatic peak detector is used to detect micro-movements above the newly proposed adaptive baseline threshold. The performance is validated on two FACS coded datasets: SAMM and CASME II. This objective method focuses on the movement of the 26 face regions. When comparing with the ground truth, the best result was an AUC of 0.7512 and 0.7261 on SAMM and CASME II, respectively. The results show that 3D HOG outperformed for micro-movement detection, compared to state-of-the-art feature representations: Local Binary Patterns in Three Orthogonal Planes and Histograms of Oriented Optical Flow.
\end{abstract}


%
\IEEEpeerreviewmaketitle

\section{Introduction}
%
%
%
%
\IEEEPARstart{F}{acial} expression research accelerated through the 1970s, and the modern theory on \textquoteleft basic\textquoteright \, emotions has generated more research than any other in the psychology of emotion~\cite{Ru97}. The universality of emotion by Ekman et al~\cite{Ek92,Ek04,Ek05} outlines 7 universal facial expressions: happy, sad, anger, fear, surprise, disgust and contempt. When an emotional episode is triggered, there is an impulse which may induce one or more of these expressions of emotion.

Micro-facial expressions (MFEs) occur when a person attempts to hide their true emotion. Once a facial expression begins to occur, the person attempts to stop it happening and causes a transient facial change. Micro-expressions usually occur in a high-stakes environment where hiding their feelings is required~\cite{Fr97a,Ek01,Po08}.

One of the main features that distinguishing MFEs from macro-facial expressions is the duration~\cite{Sh12}. The standard duration used for human and computer detection is 500 ms~\cite{Ya13a}, and will be used as such in this paper. Other definitions studied include micro-expressions lasting less than 250 ms~\cite{Ek69,Ek01}, less than 330 ms~\cite{Ek05} and less than half a second~\cite{Fr09b}. This attribute creates a micro-expression's subtle muscle movement characteristic.

Micro-facial expression analysis is less established and harder to implement due to being less distinct than normal facial expressions. Using facial point-based methods, where the movements of the face are tracked using points, are not able to track the subtle movements as accurately as with large movements. Feature representations such as Local Binary Patterns (LBP)~\cite{Oj96,Oj02,Zh07a}, Histogram of Oriented Gradients (HOG)~\cite{Da05} and Histograms of Oriented Optical Flow (HOOF)~\cite{Ch09}, are adopted instead. Despite the difficulties, micro-facial expression analysis has become increasingly popular in recent years due to the potential practical applications in security and interrogations~\cite{Os09,Fr09b,Fr09a} and healthcare~\cite{Ho92,Co09}. Automatic detection research is highly advantageous in real-world applications due to the detection accuracy of humans peaking around 40\%~\cite{Fr09b}.

Generally, the process of recognising normal facial expressions involves preprocessing, feature extraction and classification. Micro-expression recognition is not an exception, but the features extracted should be more descriptive due to the small movement in micro-expressions compared with normal expressions. One of the biggest problems faced by research in this area is the lack of publicly available datasets, which the success in facial expression recognition~\cite{Ka00,Lu10,Ya14d} research largely relies on. Gradually, datasets of spontaneously induced micro-expression have been developed~\cite{Li13a,Ya13b,Ya14a,Da15}, but earlier research was centred around posed datasets~\cite{Po09,Sh11}.
%

The outline of this paper is as follows. Section II discusses the background and related work to micro-movement analysis. Section III introduces 26 new regions created from the well-established FACS manual to help in facial feature localisation. Further, the process of fitting the mask to the face is discussed. Section IV discusses the process of extracting features from the proposed defined local regions. Section V summarises the micro-expression datasets used and the proposed histogram difference analysis method, which complements the use of a person's individualised baseline in determining when a micro-movement occurs using our proposed adaptive baseline threshold (ABT). Results and discussion follows in Section VI showing features validated on the SAMM dataset and CASME II. Section VII concludes this paper and discusses future work.
\section{Related Work}
\subsection{Face Regions}
Recent work on the recognition of MFEs have provided promising results on successful detection techniques, however there is room for improvement. To begin detection, current approaches follow methods of extracting local feature information of the face by splitting the face into regions.

Shreve et al.~\cite{Sh14} split the face into 4 quadrants and analyse each quarter as individual temporal sequences. The advantage of this method is that it is simple to analyse larger regions, however the information to retrieve from the areas are restricted to whether there was some form of movement in a more global area.

Another method is to split the face into a specific number of blocks~\cite{Zh07a,Ya14a,Da14,Da15}. The movement on the face is analysed locally, rather than a global representation of the whole face, and can focus on small changes in very specific temporal blocks. A disadvantage to this method is that it is computationally expensive to process the whole images as $m \times n$ blocks. It can also include features around the edge of the face, including hair, that do not relate to movement but could still effect the final feature vector.

Delaunay triangulation has also been used to form regions on just the face and can exclude hair and neck~\cite{Lu14}, however this approach can still extract areas of the face that would not be useful as a feature and adds further computational expense.

A more recent and less researched approach is to use defined regions of interest (ROIs) to correspond with one or more FACS AUs~\cite{Wa14b,Wa15b}. These regions have more focus on local parts of the face that move due to muscle activation. Some examples of ROI selection for micro-expression recognition and detection include discriminative response map fitting~\cite{Li15a}, Delaunay triangulation~\cite{Lu14} and facial landmark based region selection~\cite{Pa15}. Unfortunately, currently defined regions do not cover all AUs and miss some potentially important movements such as AU5 (Upper Lid Raiser), AU23 (Lip Tightener) and AU31 (Jaw Clencher).

\subsection{Micro-Movement Detection Methods}
Local Binary Patterns on Three Orthogonal Planes (LBP-TOP)~\cite{Zh07a} was extended by Davison et al.~\cite{Da14} to include Gaussian derivatives that improved on detecting movement and non-movement of the CASME II dataset than on LBP-TOP features alone. First and second order derivatives were used, with higher order features becoming more sensitive to noise. Results showed a highest accuracy of 92.6\%, but it did not report further statistical methods to take into account false negative or positive values.

Shreve et al.~\cite{Sh09b,Sh11,Sh14} proposed a novel solution of segmenting macro- and micro-expression frames in video sequences by calculating the strain magnitude in an optical flow field corresponding to the elastic deformation of facial skin tissue. The author's techniques of using strain magnitude is the most natural way of calculating whether a MFE has occurred as this is how a human would interpret using their visual system. First, macro-expressions are removed using the FE detection algorithm. Next, two additional criteria are added: (i) the strain magnitude has to be larger than surrounding regions and (ii) the duration of this increased strain can be no more than 1/5th second. As with others in this field, the limitation of datasets means that this paper could not complete a comprehensive evaluation of available data, and used their own USF-HD dataset, Canal-9~\cite{Vi09} dataset and found videos. None of these had a high-speed frame rate. The ground truth coding on the datasets have not been performed by trained FACS coders, therefore the reliability and consistency of knowing what is and is not an expression cannot be certain. The paper does not use any temporal methods due to the use of optical flow and spatial skin deformation, results in a 44\% false positive rate for detecting MFEs.

Moilanen et al.~\cite{Mo14} used an appearance-based feature difference analysis method that incorporates chi-squared (${\chi}^2$) distance to determine when a movement crosses a threshold and can be classed as a movement, following a more objective method that does not require machine learning. Peak detection was stated to have been used, however it is unknown whether this was automatic or manual. The datasets used are CASME~\cite{Ya13b} (split into A and B versions) and the original data from SMIC (not currently publicly accessible). For CASME-A the spotting accuracy (true positive rate) was 52\% with 30 false positives (FP), CASME-B had 66\% with 32 FP and SMIC-VIS-E achieved 71\% with 23 FP. The threshold value for peak detection was set semi-automatically, with a percentage value between [0,1] being manually set for each dataset. Only spatial appearance is used for descriptor calculation, therefore leaving out temporal planes associated with video volumes.

By exploiting the feature difference contrast, Li et al.~\cite{Li15b} proposed an algorithm that spots micro-expressions in videos. Further, they combine this with an automatic micro-expression analysis system to recognise expressions of the SMIC-E-VIS dataset in one of three categories: positive, negative or surprise. This spotting algorithm is very similar to the feature difference method seen in~\cite{Mo14}, however HOOF is compared along with LBP. The highest result came from using LBP in the CASME II dataset with an AUC of 92.98\%. The best performance came from using the SMIC-E-VIS dataset and LBP, where the spotting accuracy was about 70\% of micro-expressions, 13.5\% FPR, and the AUC was 84.53\%.

The proposed micro-expression spotting system in~\cite{Li15b} still uses a block-based approach, as in~\cite{Mo14}, therefore potentially included redundant information. There is also no face alignment performed for these video clips which could lead to head movements being falsely spotted. It should also be noted that throughout this paper, the use of \textquoteleft accuracy\textquoteright \, is not defined. Using more reliable statistical measures, such as recall, precision and F-measure, would allow the system to be scrutinised more effectively.

A Main Direction Mean Optical Flow (MDMO) feature was proposed by Liu et al.~\cite{Li15a} for micro-facial expression recognition using SVM as a classifier. The feature is based around histogram of oriented optical flow (HOOF)~\cite{Ch09}. The method of detection also uses 36 regions, partitioned using 66 facial points on the face, to isolate local areas for analysis, but keeping the feature vector small for computational efficiency. The best result on the CASME II dataset was 67.37\% using leave-one-subject-out cross validation.

A further optical flow-based method was proposed by Xu et al.~\cite{Xu16} by performing pixel-level alignment for each micro-expression sequence. This facial dynamics map represented micro-expression movement better with principal directions. Using the fine alignment, the results are improved compared to without this step, however the time taken to process features is still relatively large due to increased complexity and optical flow calculations on a micro-expression sequence.

Block-based approaches, proposed in~\cite{Da15,Mo14,Li15b}, split the face into $m\times n$ blocks. Doing this can include a lot of redundant information, in other words, non-muscle movement. In addition, there are no indication of  where on the face the movement occurs. 

A better solution is required to focus on areas of the face that provide important information and to test micro-movement feature descriptors on different datasets. Previously determined thresholds on when a micro-movement occur are set using the data provided from the movements themselves~\cite{Mo14}. To solve this would require the use of a person's baseline expression where they show no evidence of facial muscle movement. Setting a threshold based on this value would allow for an adaptive approach in selecting when a micro-movement has occurred.
%

%
\section{Proposed FACS-Based Regions}
\begin{figure}
\centering
\includegraphics[scale=0.256]{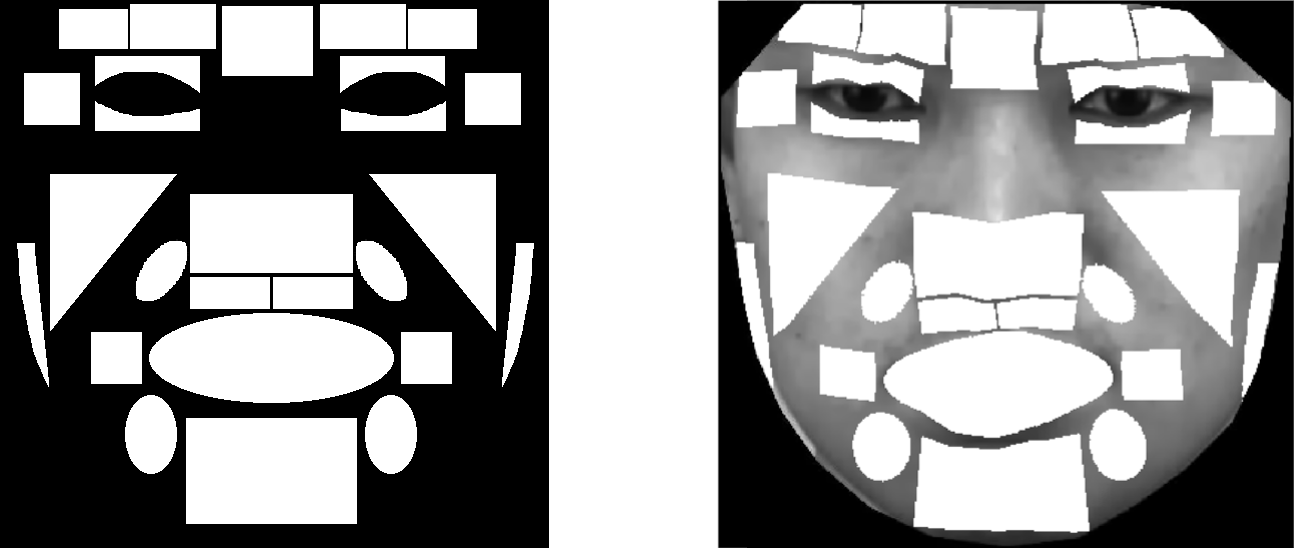}
\caption{The overall binary mask with 26 regions (left) and the same regions fitted to an aligned and warped frame (right).}
\label{fig:maskFit}
\end{figure}

To ensure only relevant movements are detected, 26 FACS-based~\cite{Ek78a} regions are proposed. Each region has been selected by three FACS-certified coders to focus only on areas of the face that contain movements from particular AUs. The advantage of this is that features can be locally analysed, the intensity of individual regions can be independently studied and not processing insignificant parts of the face (e.g. the hair). Fig.~\ref{fig:maskFit} shows the overall binary mask used to select regions of the face and an example of the mask being fitted to a warped frame.

In addition to the proposed regions for micro-movement analysis, we outline the process of aligning faces and using piecewise affine warping~\cite{Co04} to warp the regions to fit the facial features of a person using automatically detected facial feature points.

\subsection{Facial Point Detection and Alignment}
Under a controlled experimental setting, most participants sit relatively still and keep head movements to a minimum. Regardless of this, a lot of facial analysis methods align all the faces within the dataset to a canonical pose for examination. One method is to use facial points detected on the face, and then warp the face while preserving the facial features for analysis. This method will be used in this paper for alignment of temporal movements by automatically detecting the face points and warping the faces in each sequence to common points using piecewise affine warping~\cite{Co04}.

\subsubsection{Automatic Face Point Detection}
To detect the facial points automatically, the Face++ facial point detection algorithm~\cite{Me13b} was used. For temporal sequences, 83 points were detected (see Fig.~\ref{fig:faceWarp}(b)) in the first frame of the movement sequence and used as control points for that face. As the movements sequences are very short in length, there is little head movement to distort the warping in subsequent frames. A similar process was used when detecting the face points in the CASME and CASME II datasets~\cite{Ya13b,Ya14a}.

\subsubsection{Face Alignment}
To reduce the amount of rigid head movement that is inevitable when recording humans for long periods of time, each video frame of the micro-movement videos are aligned to be registered with the first frame of the sequence. As the micro-movements are subtle and last no more than 100 frames, the first frame is used as a reference frame.

To complete the alignment on our images without the need for facial points we use the method proposed by Guizar-Sicairos et al.~\cite{Gu08}. This method uses a 2D-Discrete Fourier Transform (2D-DFT) to calculate the cross-correlation and find its peak, which can then be used to find a translation between the reference image, $f(x,y)$ and image to register, $g(x,y)$.

The first step is to apply the 2D-DFT to both images. To obtain subpixel accuracy, the resolution of $f(x,y)$ is upsampled from the image dimensions of $M$ by $N$ to $k*M$ by $k*N$, where $k$ is an upsampling factor that defines a subpixel error of $1/k$. For the proposed method, $k$ is set to 100.

An initial estimate of the translation to be applied to $g(x,y)$ is defined by $T(x,y)$ and is calculated by finding the cross-correlation peak using the inverse 2D-DFT and setting $k$ to be 2. The 2D-DFT of the images are embedded into an array that is twice the size of the original image. The images can then be aligned to the estimated $T(x,y)$. This process continues until an upsampling factor of $k$ is achieved.

By using a 2D-DFT, this method is able to select a local neighbourhood, around the initial peak estimate, to find the final peak calculated from cross-correlation rather wasting computing resources by calculating over the entire upsampled array. When using such large amount of images from high-speed video, processing efficiency is highly valued and provides a further step towards real-time micro-movement analysis.
\subsection{Piecewise Affine Warping}
If we require an image $\textbf{I}$ to be warped to a new image $\textbf{I}'$, $n$ control points, $x_{i}$, are mapped to new points $x'_{i}$. To understand the process of image warping, Cootes and Taylor~\cite{Co04} describe the continuous vector valued mapping function, $\textbf{f}$, to project each pixel of image $\textbf{I}$ to the new image $\textbf{I}'$. The mapping function $\textbf{f}$ is defined as
\begin{equation}
\textbf{f}(x_{i}) = x'_{i}\; \, \, \, \forall i = 1,\ldots,n
\label{eq:mapFunc}
\end{equation}
To avoid holes and interpolation issues, it is better to calculate the reverse mapping, $\textbf{f}'$ taking the points of $x'_{i}$ into $x_{i}$. For each pixel in the newly warped image, $\textbf{I}'$, it can be determined where in the original image, $\textbf{I}$, it came from and fill it in. The mapping function $\textbf{f}$ can also be broken down into a sum,
\begin{equation}
\textbf{f}(x) = \sum\limits_{i=1}^{n} f_{i}(x)x'_{i}
\label{eq:mapFuncSum}
\end{equation}
\noindent where the $n$ continuous scalar valued functions $f_{i}$ each satisfy
\begin{equation}
f_{i}(x_{j})=
\begin{cases}
1, & \text{if}\ i=j \\
0, & \text{otherwise}
\end{cases}
\label{eq:funcSatisfy}
\end{equation}
ensuring that $\textbf{f}(x_{i}) = x'_{i}$.

\begin{figure*}
\centering
\includegraphics[scale=0.166]{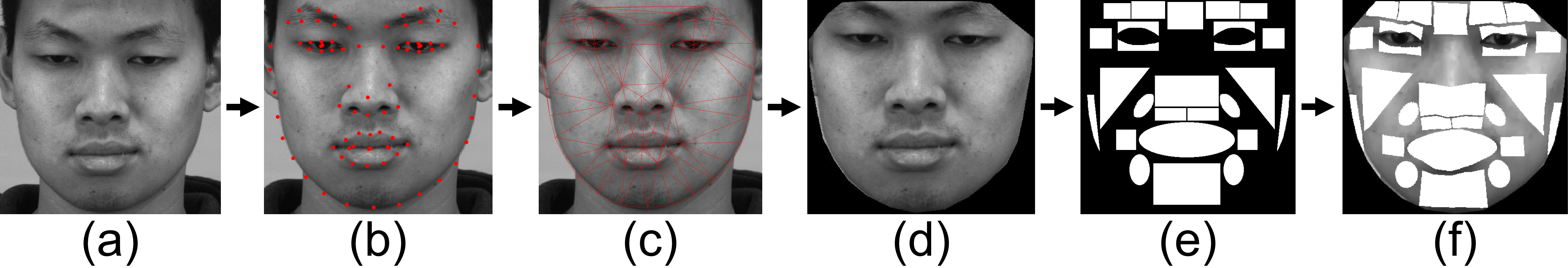}
\caption{The process of piecewise affine warping. (a) the original image (b) the original face has points automatically detected (c) Delaunay triangulation creates the convex hull for to allow the mask to be warped (d) the cropped face used for feature extraction and mask fitting (e) the original mask is warped to the shape of the cropped face (f) the final warped mask that has been applied to the cropped face.}
\label{fig:faceWarp}
\end{figure*}

One of the simplest forms of warping function is piece-wise affine (PWA)~\cite{Co04}, where each $f_{i}$ is assumed to be linear in a local region and zero everywhere else. As an example, in the one dimensional case (where each $\textbf{x}$ is a point on a line), suppose the control points are arranged in ascending order ($x_{i}<x_{i+1}$). To arrange $\textbf{f}$ so that it will map a point $\textbf{x}$, that is halfway between $x_{i}$ and $x_{i+1}$, to a point halfway between $x'_{i}$ and $x'_{i+1}$, the following setting is applied
\begin{equation}
f_{i}(x)=
\begin{cases}
(x - x_{i})/(x_{i+1} - x_{i}), & \text{if}\ x\in [x_{i},x_{i+1}]\ \text{\&}\ i<n \\
(x - x_{i})/(x_{i} - x_{i-1}), & \text{if}\ x\in [x_{i-1},x_{i}]\ \text{\&}\ i>1 \\
0, & \text{otherwise}
\end{cases}
\label{eq:funcSetting}
\end{equation}
\noindent where the control points in the region is between $x_{1}$ and $x_{n}$ and the image can only be warped between these points.
As this paper is warping images in 2-dimensions (2D), triangulation using the Delaunay method is used to partition the convex hull of the control points into a set of triangles. For the points within each triangle, an affine transformation is applied which uniquely maps the corners of the triangle to their new positions in $\textbf{I}'$ (see Fig.\ref{fig:faceWarp} for an example of a single frame being warped).
To illustrate is transformation, suppose $\textbf{x}_{1}$, $\textbf{x}_{2}$ and $\textbf{x}_{3}$ are the three vertices of a triangle within the convex hull. Any internal point of the triangle can be written as
\begin{equation}
\begin{split}
\textbf{x}&= \textbf{x}_{1} + \beta(\textbf{x}_{2} - \textbf{x}_{1}) + \gamma(\textbf{x}_{3} - \textbf{x}_{1})\\
&= \alpha\textbf{x}_{1} + \beta\textbf{x}_{2} + \gamma\textbf{x}_{3}
\end{split}
\label{eq:inTriangle}
\end{equation}
\noindent where $\alpha = 1 - (\beta + \gamma)$ and so $\alpha + \beta + \gamma = 1$. For $\textbf{x}$ to be inside the triangle, $0\leq\alpha,\beta,\gamma\leq 1$. Under the affine transformation, this point maps to
\begin{equation}
\textbf{x}'= \textbf{f}(\textbf{x}) = \alpha\textbf{x}'_{1} + \beta\textbf{x}'_{2} + \gamma\textbf{x}'_{3}
\label{eq:affinePoint}
\end{equation}
To create the warped image, each pixel $\textbf{x}'$ in $\textbf{I}'$ is set to the triangle it belongs to, the $\alpha, \beta$ and $\gamma$ coefficients are computed to give the pixel's relative position in the triangle, which are then used to find the equivalent point in the original image, $\textbf{I}$. The point is sampled and the value copied into pixel $\textbf{x}'$ in $\textbf{I}'$.

Although the PWA transformation gives continuous deformation, it is not smooth. This leads to straight lines being kinked across the triangle boundaries. Fortunately, face textures have few straight lines and are minimally effected by this issue. As every person has a different face shape, PWA transformation can generalise this shape while keeping the person's individual features and allows for consistent region analysis discussed in the next Section.

\subsection{FACS-Based Region Fitting}
In the representation of a binary mask shown in Fig.~\ref{fig:faceWarp}, each region is warped so that they fit to the correct area of the face. The other option is to warp the face to a static mask shape, however this leads to faces becoming distorted and potentially removing any micro-movements. One example of a problem arising by warping the face to the mask is that regions may not fit to the best position. This includes when a person is wearing glasses or have facial features that are much different to usual, such as a very large nose.

Using the FACS-based regions defined, each participant has a mask individually warped to their faces based on the points obtained during PWA. Even though the mask changes shape based on the person's facial features this does not affect movement classification as each participant in a dataset is handled individually rather than a set of abstract classes. When warping the mask to the individual face points, the different face proportions are taken into account to avoid personal face differences.

After aligning and warping the facial movements, feature extraction can be applied to whole image. Using specific parts of the face can provide a solution, as it focuses on the important local parts of the face responsible for expressions.
\section{Micro-Movement Region Localisation}
\subsection{Spatial and Temporal Plane Analysis}
\begin{figure}
	\centering
	\includegraphics[scale=0.16]{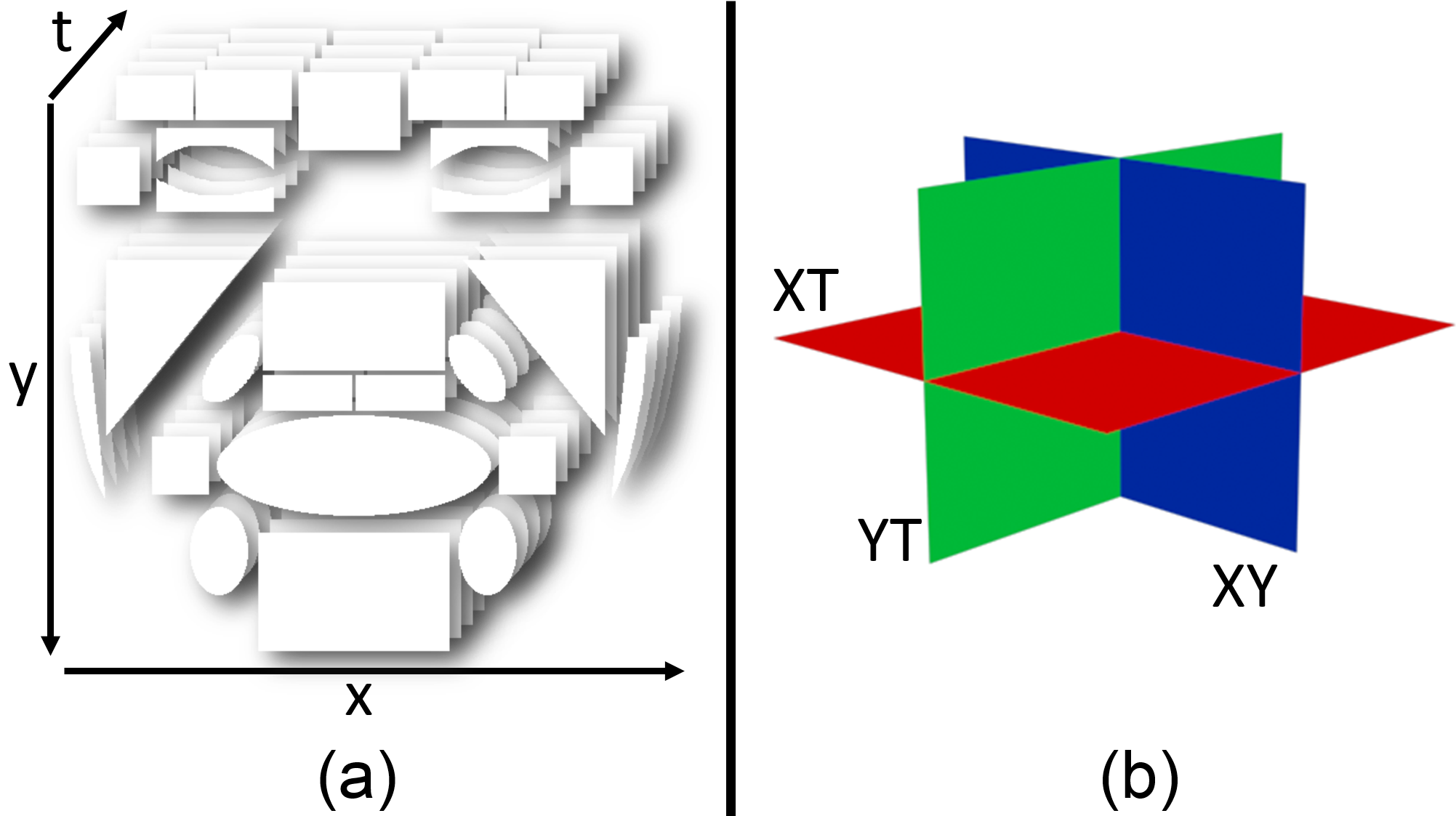}
	\caption{(a) Visual representation of the spatio-temporal configuration of video frames split into blocks. (b) The XY, XT, YT planes used for feature analysis in LBP-TOP and 3D HOG.}
	\label{fig:blocksAndPlanes}
\end{figure}

HOG originally derives from a spatial planes only~\cite{Da05}, with success in recognising people. It is only natural to use a spatio-temporal feature when analysing video data. Fig.~\ref{fig:blocksAndPlanes}(a) shows regions as a \textquoteleft video cube\textquoteright \, that contain each frame of the video cropped in that particular region. Fig.~\ref{fig:blocksAndPlanes}(b) shows a representation of three orthogonal planes in 3D space. Each plane outlines the axis which is being analysed in the video - XY, XT and YT.
\subsection{Feature Extraction}
For the first time in micro-movement detection, 3D HOG is used as a feature descriptor for video sequences. Initial processing includes de-noising to reduce high-speed video noise. Three planes (XY, XT and YT) are extracted using 3D HOG to describe different directions of motion. Two additional plane representations are created by concatenating the XY, XT and YT planes and the temporal XT and YT planes.

From both datasets, feature extraction is used to represent the movements and baselines defined by the FACS coding ground truth. Each movement has the 3D HOG feature applied to extract the temporal feature representation in each of the defined regions. The extracted movements also include neutral frames (200 in SAMM~\cite{Da16} and 50 in CASME II~\cite{Ya14a}) at the start and end of the sequence so the movement has an onset and offset point like a natural expression would. The start and end point of the movements have been FACS coded for both datasets.

The baseline feature is extracted similarly to the movements, but uses a set amount of frames to define a subject's baseline. The features are once again split into regions to represent the baseline of that particular local area for that particular person.

\subsection{Temporal De-Noising}
All videos captured can contain some form of noise due to the way images are captured digitally, whether this be through lighting, temperature or equipment malfunction. High-speed video is particularly susceptible due to capturing lots of images in a short space of time. To counter this, de-noising is applied using a sparse signal processing method~\cite{Da07} called collaborative filtering. This processes the video volume block-wise and attenuates noise to reveal fine details shared by the block groups while preserving the unique features of each block. The output is the same size frames of the video, so no further processing is required to suit the feature descriptors.
\begin{figure*}
\centering
\includegraphics[scale=0.22]{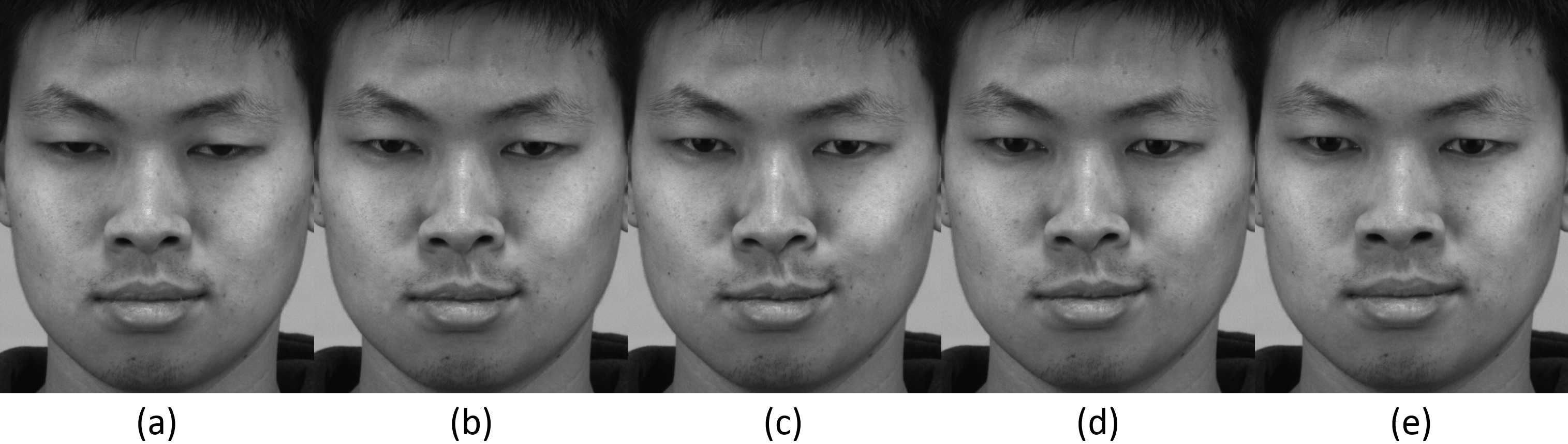}
\caption {An example of a coded micro-movement. The movement is AU 13 and AU 39, which is the sharp lid puller and nostrils compressing. Image (a) is the onset frame, (c) is the apex where the mouth curls upwards sharply and the nostrils move inwards. Finally, (e) is the offset frame.}
\label{fig:sammExample}
\end{figure*}
\section{Micro-Facial Movement Detection}
Current literature focuses on the recognition of micro-expressions based under emotional contexts~\cite{Ya13b,Ya14a,Pf11a,Da14}. Unfortunately by doing this assumes the pattern of micro-facial movements are similar to their macro-expression counterparts, for example if the mouth moves in a micro-expression, this means happiness. These methods also rely on machines learning to model and then classify each type of micro-facial expression. In a real-world scenario, micro-movements could manifest in a variety of ways, and it is near impossible to split these movements into distinct classes.

In the proposed method, micro-movements are treated objectively as facial muscle activations, like in FACS, and the temporal differences are calculated using the $\chi^{2}$ distance. Previous difference methods~\cite{Mo14,Sh11,Sh14} all assume that the detected movements in the videos are actual micro-expressions. The datasets they use can sometimes confirm this, but it creates a very constrained analysis. The proposed method uses a baseline expression vector calculated from the neutral face sequence of participants to create individual baseline thresholds to determine what is a micro-movement for that person.

The distance function described derives from Moilanen et al.~\cite{Mo14} but extends to the temporal domain, uses FACS-based regions with 3D HOG, region normalisation and baseline thresholding. Finally, peak detection is used to automatically find movement peaks and output the onset, apex and offset frames of each movement.
\subsection{Micro-Movement Datasets}
Datasets that contain spontaneous micro-facial movements or expressions are limited~\cite{Li13a,Ya13b,Ya14a}, but are required to effectively analyse these movements. Posed micro-facial expression datasets exist~\cite{Po09,Sh11} but are not representative of a genuine micro-movement. In response to the lack of spontaneous datasets, we use the SAMM~\cite{Da16} dataset alongside the CASME II dataset to validate our proposed method.

The SMIC dataset~\cite{Li13a} is one of the first datasets to induce spontaneous micro-movements, shortly followed by the original CASME dataset~\cite{Ya13b}. However, the SMIC and CASME datasets have a lower frame rate of 100 and 60 fps respectively. The SMIC also does not provide FACS coding. This leaves the CASME II dataset, which provides the fairest comparison with SAMM.

\subsubsection{SAMM}
SAMM~\cite{Da16} is used in the experiments and contains 159 micro-movements captured at 200 frames per second (fps). To obtain a wide variety of emotional responses, the dataset was required to be as diverse as possible. A total of 32 participants were recruited for the experiment from within the university with a mean age of 33.24 years (standard deviation: 11.32, ages between 19 and 57). The ethnicities of the participants are 17 White British, 3 Chinese, 2 Arab, 2 Malay and 1 each: African, Afro-Caribbean, Black British, White British / Arab, Indian, Nepalese, Pakistani and Spanish. An even gender split was also achieved, with 16 male and 16 female participants.

Rather than making inferences about the emotional context, for the SAMM dataset treats all movements as objective, with no assumptions made about the emotion after each experimental stimulus. The dataset was FACS coded by three certified coders, achieving an inter-reliability score of 0.82. Coding was performed after the videos were recorded as per usual FACS coding procedure. Every movement is coded, including the macro-movement, with an onset, apex and offset frame recorded to note the duration. Unlike most other datasets, every FACS AU is coded regardless of their relation to emotions. This includes head and eye movement codes. By FACS coding the data comprehensively, the dataset can be used for a much wider purpose when it becomes publicly available. An example of a micro-movement is as illustrated in Fig.~\ref{fig:sammExample}. Baselines of people are required for our method, and the SAMM dataset contains large baseline sequences for each participant to allow for individualised analysis. The amount of neutral frames used for SAMM was 200 before and after the movement clip.

The SAMM dataset is available online for download (http://goo.gl/SJmoti) for research purposes and contains 4GB of images formatted to jpeg.
\subsubsection{CASME II}
CASME II~\cite{Ya14a} was created as an extension of the original CASME dataset~\cite{Ya13b}. The frame rate increased to 200 fps to analyse more detail in muscle movements, and 247 newly FACS coded MFEs from 26 participants were obtained. The facial area used for analysis was the larger than CASME and SMIC at 280$\times$340 pixels. However, as with CASME, this dataset includes only Chinese participants and categorises in the same way. Both CASME and CASME II used 35 participants, mostly students with a mean age of 22.03 (SD = 1.60). Along with only using one ethnicity, both these datasets have the disadvantage of using young participants only, restricting the dataset to analysing similar looking participants (based on age features).

Unlike SAMM, the CASME II dataset does not provide explicit baseline data, therefore excess frames before and after the annotated movements, provided by the CASME II dataset, are used for baseline calculations. Ideally, all micro-movement datasets created in the future will provide baseline sequences. The amount of neutral frames used for movement sequences was 50 for CASME II due to the lack of available neutral frames.
\subsection{Histogram Distance Analysis}
3D HOG can be described as a histogram-based feature, and so is perfect for using a feature distance measure to calculate the changes between frames. The concept of feature difference analysis is for each current frame being processed, it is compared with the average feature frame that is represented by the average features of the start frame, \textit{k}-th frame before the current frame, and the end frame, \textit{k}-th frame after the current frame. The \textit{k}-th frame is described as
\begin{equation}
\textit{k} = \frac{1}{2}(\textit{N} - 1)
\label{eq:kFrame}
\end{equation}
\noindent where \textit{N} is the micro-interval value that is always set to an odd number. Moilanen et al.~\cite{Mo14} sets the micro-interval to 21 for the CASME~\cite{Ya13b} dataset that was recorded at 60 fps. As the SAMM dataset was recorded at 200 fps~\cite{Da16}, it is calculated that the value of \textit{N} should be around 71. The value of \textit{k} in this instance would be 35, similar to the settings in~\cite{Da15}.

The difference between the current frame and average feature frame shows the facial changes in a particular region and the possible change in the features is rapid since it occurs between start frame and end frame, to distinguish the quick changes from temporally longer events. The difference analysis continues for each frame except the first and last \textit{k} frames that would exceed the boundaries of the video. This also means rapid facial movements such as blinks would also be classified as a movement, and so our mask removes as much of the eye as possible without removing the important muscle areas around the eyes.

The difference that is calculated can also be described as the dissimilarity between histograms in each region. The $\chi^{2}$ distance is such as measure and is defined as
\begin{equation}
\chi^{2}(P,Q) = \sum_{b}^{B}\frac{(P_{b} - Q_{b})^{2}} {(P_{b} + Q_{b})}
\label{eq:chisquare}
\end{equation}
\noindent where $b$ is the $b$-th bin in the $P$ and $Q$ histograms that have an equal number of bins for a total amount of bins $B$. This equation can be used in all temporal planes (XY, XT, YT) to calculate dissimilarity. $\chi^{2}$ has been applied successfully to applications such as object and text classification~\cite{Zh07b}.

In many other histogram distance measures, the differences between large bin values are less significant than between small bins~\cite{Pe10}, however as this is calculating dissimilarity the difference should not be dependent on the scale of the bin values when each feature is deemed equally important. Ahonen et al.~\cite{Ah06} found that the $\chi^{2}$ distance performs better in face analysis task than histogram intersection or log-likelihood distance.

Unlike previous methods that split the images into even-sized blocks, the regions used in this method are all different sizes. This leads to the issue of the larger regions having more pixels in the area of the mask and therefore differences that are not reflective of which regions is more meaningful. The normalisation is defined as
\begin{equation}
\textbf{F}_{i} = \frac{\textbf{F}_{i}} {A_{r}}
\label{eq:regionNorm}
\end{equation}
\noindent where $A_{r}$ is the number of pixels in each individual binary bask region and $i$ is the index of the value within the feature $\textbf{F}$. This step is completed for each region and was done to make sure that the area of regions that were not rectangular in a matrix were still accurately calculated to normalise the regions correctly.

There is an option to select the top regions, defined as $\textit{R}$, that are used to rank the greatest difference values for each region. For example, if $\textit{R}$ was set to 8, then the 8 regions with the greatest difference values would used to form the first feature vector, $\textbf{F}_{\textit{R},i}$, that represents the overall movement of the face. It is defined as
\begin{equation}
\textbf{F}_{\textit{R},i} = \sum_{r = 1}^{\textit{R}} (D_{r,1}, D_{r,2},\ldots, D_{r,i})
\label{eq:initialFeat}
\end{equation}
\noindent where $D$ is the difference values of each individual region, $j$, sorted in descending order up to $\textit{R}$ for each frame, with $i$ being the total number of frames or index value of that feature. For the proposed method, each region feature is calculated initially and ranked in descending order from highest difference value to lowest. By sampling $\textit{R}$ regions (2, 4, 6, ... , 26), we produce Receiver Operating Characteristic (ROC) curves and the Area Under Curve (AUC) is calculated.

To handle the noise from local magnitude variations and video artefacts, the average values of the frames $i+k$ and $i-k$ are subtracted from the value in the current frame. Each new value of the $i$-th feature of region $\textit{R}$ is therefore calculated as
\begin{equation}
\textbf{F}_{\textit{R},i}' = \textbf{F}_{\textit{R},i} - \frac{1}{2} (\textbf{F}_{\textit{R},i+k} + \textbf{F}_{\textit{R},i-k} )
\label{eq:secondFeat}
\end{equation}
\noindent with each frame having Eq.~\ref{eq:secondFeat} applied, apart from the first and last \textit{k} frames of the video sequence due to the temporal boundaries. Finally, any negative values in $\textbf{F}_{\textit{R},i}'$ are set to zero. The reason for this is that any values lower than zero indicates that the value at the current frame was below the average difference values of the frame $i+k$ and $i-k$. Therefore, there are no fast changes in current frame.
\subsection{Individualised Baselines}
In any emotional environment where a person is observed for deception or concealment of true feelings, it is important to know their baseline so one does not get confused with a facial twitch or movement that a person consistently performs. As this is common practice when analysing humans from a Psychological perspective~\cite{Wa81}, there is a strong reason to believe using baseline features in computer vision would produce good results.

The original threshold method by Moilanen et al.~\cite{Mo14} relied on taking the mean and maximum values of the final feature vector to detect a valid peak. Unfortunately, this meant that a peak will always be \textquoteleft detected\textquoteright \, regardless of there being a real movement or not. The newly designed threshold is based on the neutral expression sequence from each individual subject in the dataset. This sequence was processed the same way as the movements, but by using an ABT, we take into account the mean of both the movement and baseline feature vector that adapts the threshold level based on a balance between what is happening on the participant's face and what their baseline expression level is. The ABT can be calculated by
\begin{equation}
ABT =
\begin{cases}
max(\varrho), & \text{if}\ max(\varrho) > \bar{\epsilon} \\
\frac{\bar{\epsilon} + \bar{\varrho}}{2}, & otherwise \\
\end{cases}
\label{eq:ABT}
\end{equation}
\noindent where $ABT$ is the calculated adaptive threshold, $\varrho$ is the baseline feature vector and $\bar{\varrho}$ is its mean. The movement feature vector and its mean is denoted by $\epsilon$ and $\bar{\epsilon}$ respectively.
\subsection{Peak Detection and Temporal Phase Identification}
The peak outputs are easily seen as small Gaussian shaped curves when visualised. Using publicly available MATLAB code~\cite{Ha15}, the peaks are detected using signal processing methods. The first derivative of the signal is smoothed, and then peaks are found from the downward-going zero-crossings. Then only the zero-crossings whose slope exceeds a defined \textquoteleft slope threshold\textquoteright \, at a point where the original signal exceeds a certain height or \textquoteleft amplitude threshold\textquoteright.
\subsubsection{Temporal Phases of Peaks}
Estimating the apex simply uses the highest point of the found Gaussian curves. Estimating the \textquoteleft start\textquoteright \, and \textquoteleft end\textquoteright \, of the peak (the onset and offset value respectively), is a bit more arbitrary, because typical peak shapes approach the baseline asymptotically far from the peak maximum. Peak start and end points can be set to around 1\% of the peak height, but any random noise on the points during peak detection will often be a large fraction of the signal amplitude at that point. 

Smoothing is done within this process to reduce noise, however this can lead to distortion of peak shapes and change the start and end points. One solution is to fit each peak to a model shape, then calculate the peak start and end from the model expression. This minimises the noise problem by fitting the data over the entire peak, but it works only if the peaks can be accurately modelled. For example, Gaussian peaks reach a fraction $a$ of the peak height of 
\begin{equation}
x = p \pm\sqrt{\left(\frac{w^{2}\log(\frac{1}{a})} {2\sqrt (\log(2))}\right)}
\label{eq:peakStartEnd}
\end{equation}
where $p$ is the peak position and $w$ is the peak width~\cite{Ha15}. So if $a=0.01$, $x=p\pm1.288784\times w$.
\section{Results and Discussion}
\begin{table}
\centering
\renewcommand{\arraystretch}{1.3}
\caption{The AUC values for each feature descriptor on both datasets.}
\label{tab:AUC}
\begin{tabular}{|l|l||l|}
\hline
Feature & SAMM & CASME II \\ \hline
3D HOG - XY & 0.6910 & 0.6454 \\ \hline
3D HOG - XT & \textbf{0.7513} & \textbf{0.7261} \\ \hline
3D HOG - YT & 0.7278 & 0.6486 \\ \hline
3D HOG - XTYT & 0.7513 & 0.7157 \\ \hline
3D HOG - All Planes & 0.7481 & 0.7183 \\ \hline \hline
LBP-TOP - XY & \textbf{0.6401} & 0.4631 \\ \hline
LBP-TOP - XT & 0.6197 & 0.4853 \\ \hline
LBP-TOP - YT & 0.6173 & \textbf{0.4866} \\ \hline
LBP-TOP - XTYT & 0.6144 & 0.4747 \\ \hline
LBP-TOP - All Planes & 0.6154 & 0.4623 \\ \hline \hline
HOOF & \textbf{0.6312} & \textbf{0.5203} \\ \hline
\end{tabular}
\end{table}
The proposed method of using FACS-based regions with individualised baselines performs well on both SAMM and CASME II compared with the previous state of the art. Two other feature descriptors that have been used in micro-movement detection are implemented to compare with 3D HOG.

We performed a peak spotting check that determined whether the apex of a detected peak was above the ABT and within the duration of the micro-movement that had been labelled during FACS coding. The ground truth was set for each region, where the ground truth AUs corresponded to regions of movement. With each movement containing the ground truth for every region, it was possible to extract the common true positives (TP), false positives (FP), false negatives (FN) and the less common true negatives (TN).
\begin{figure}
\centering
\includegraphics[scale=0.59]{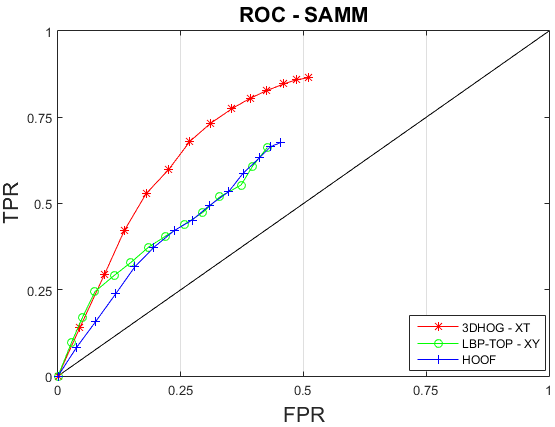}
\caption{Three ROC curves are shown for the SAMM dataset. Each descriptor corresponds to a different curve with 3D HOG performing best.}
\label{fig:rocSAMM}
\end{figure}
\begin{figure}
\centering
\includegraphics[scale=0.59]{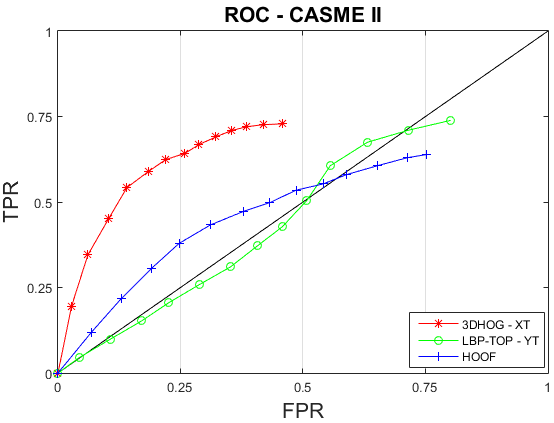}
\caption{Three ROC curves are shown for the CASME II dataset. Each descriptor corresponds to a different curve with 3D HOG performing best.}
\label{fig:rocCASME}
\end{figure}
\begin{table*}
	\centering
	\renewcommand{\arraystretch}{1.3}
	\caption{Performance analysis metrics for each feature descriptor on both datasets. The results displayed show the best performing chosen top regions which was found to be 12.}
	\label{tab:results}
\begin{tabular}{|l|l|l|l|l||l|l|l|l|}
\hline
 & \multicolumn{4}{c||}{SAMM} & \multicolumn{4}{c|}{CASME II} \\ \hline
Feature & \multicolumn{1}{c|}{Recall} & \multicolumn{1}{c|}{Precision} & \multicolumn{1}{c|}{F-Measure} & \multicolumn{1}{c||}{Accuracy (\%)} & \multicolumn{1}{c|}{Recall} & \multicolumn{1}{c|}{Precision} & \multicolumn{1}{c|}{F-Measure} & \multicolumn{1}{c|}{Accuracy (\%)} \\ \hline
3D HOG - XY & 0.5607 & 0.2831 & 0.3763 & 70.87 & 0.5134 & 0.4216 & 0.4631 & 65.77 \\ \hline
3D HOG - XT & \textbf{0.6804} & 0.3198 & 0.4352 & 72.32 & 0.6235 & \textbf{0.5341} & \textbf{0.5754} & 73.55 \\ \hline
3D HOG - YT & 0.5821 & 0.3043 & 0.3998 & 72.60 & 0.4857 & 0.4561 & 0.4705 & 68.57 \\ \hline
3D HOG - XTYT & 0.6661 & 0.3179 & 0.4305 & 72.38 & 0.6202 & 0.5157 & 0.5631 & 72.34 \\ \hline
3D HOG - All Planes & 0.6607 & 0.3181 & 0.4295 & 72.49 & 0.6202 & 0.5164 & 0.5636 & 72.39 \\ \hline
LBP-TOP - XY & 0.3732 & 0.2735 & 0.3157 & 74.65 & 0.2807 & 0.2427 & 0.2603 & 54.15 \\ \hline \hline
LBP-TOP - XT & 0.35 & 0.2688 & 0.3041 & 74.90 & 0.3513 & 0.2876 & 0.3163 & 56.35 \\ \hline
LBP-TOP - YT & 0.3196 & 0.2716 & 0.2937 & 75.90 & 0.3101 & 0.2622 & 0.2842 & 55.09 \\ \hline
LBP-TOP - XTYT & 0.3464 & 0.2755 & 0.3070 & 75.48 & 0.3319 & 0.2714 & 0.2987 & 55.19 \\ \hline
LBP-TOP - All Planes & 0.3482 & 0.2589 & 0.2970 & 74.17 & 0.3134 & 0.2560 & 0.2818 & 54.08 \\ \hline \hline
HOOF & 0.4214 & 0.2486 & 0.3128 & 70.98 & 0.4723 & 0.3351 & 0.3920 & 57.90 \\ \hline
\end{tabular}
\end{table*}
Firstly we present the ROC curves for three types of feature descriptors on the SAMM (Fig.~\ref{fig:rocSAMM}) and CASME II (Fig.~\ref{fig:rocCASME}) that plots the true positive rate (TPR) on the y-axis and the false positive rate (FPR) on the x-axis. The TPR or recall is calculated by $\frac{TP}{(TP+FN)}$ and FPR is calculated by $\frac{FP}{(FP+TN)}$. The points of the ROC curves are calculated by iterating from the 2 regions with the highest difference values for the feature descriptor up until all regions are used. As the curves show, as the more regions are introduced, both TPR and FPR increases as predicted when more regions become available for detection. Further, the plots show that 3D HOG outperforms LBP-TOP and HOOF on both datasets.

In addition to the ROC curves, we calculated the area under curve (AUC) for each feature in both datasets in Table~\ref{tab:AUC}. 3D HOG performs best on the SAMM and CASME II dataset in the XT plane with 0.7513 and 0.7261 respectively. LBP-TOP achieves 0.6401 in the XY planes for the SAMM dataset, however is does not perform as well in the CASME II dataset with the highest result in the YT plane with 0.4866.

We present the performance analysis scores of recall (TPR), precision and F-measure in Table~\ref{tab:results}. Each score was selected from the number of \textquoteleft top\textquoteright \, regions that performed best in describing the micro-movements, which in these experiments was 12 regions. Overall the best feature descriptor for SAMM was 3D HOG in the XT plane achieving 0.6804, 0.5341 and 0.5754 for recall, precision and F-measure respectively.

In regards to spotting accuracy, our method performs well and is comparable with the results in~\cite{Mo14,Li15b}. However the precision is low on the majority of cases due to the larger amount of false positives detected. Unfortunately, as with any recordings of humans in video sequences, it is difficult to eliminate all head movements, even with face alignment. As micro-movements are subtle motions, any head movements are likely to contribute to the final difference values, therefore skewing the accuracy of spotting real micro-movements.

The higher precision seen in CASME II compared with SAMM can be explained by the lower number of frames that were available to analyse in the CASME II dataset, meaning false positives are less likely and spotting is easier. The original recordings of the CASME II participants are no longer available, and so creating a longer sequence was not possible at this time.

Despite using only spontaneous datasets, the data collection required the participants to stay still to minimise head movements under a controlled environment, meaning the method can be influenced by large head movements. In an unconstrained situation, normal human movement would be problematic, however inducing micro-expressions in the wild is almost impossible.

Higher spotting accuracy can be achieved through the increase of regions, however this tips the balance of TPs and FPs, lowering the overall F-measure. For example, using the 12 regions described in Table~\ref{tab:results} the 3D HOG - XT feature on SAMM the recall is 0.6804, however if this is increased to 26 regions the recall score jumps to 0.8643 but is accompanied by a lower F-measure of 0.3752. The use of ranking regions to balance the trade off between detecting more and less sensitivity was proposed in Moilanen et al.~\cite{Mo14} and advanced upon by Davison et al.~\cite{Da15} to remove features that may introduce noise and redundant information.
\section{Conclusion}
This paper has shown evidence for the use of neutral expression baseline sequences as a comparison to detect micro-facial movements using FACS-based regions. Both 3D HOG and LBP-TOP perform well, however 3D HOG performed the best with a recall of 0.6804 and AUC of 0.7513. This method takes away any assumption based on emotions and reduces the movements to their basic definition of muscle movements. Doing this allows for further interpretation at a later stage rather than attempting to use techniques, such as machine learning, to define micro-movements into discrete categories.

The proposed FACS-based regions follows a recent approach that moves away from blocking methods~\cite{Lu14,Li15a,Pa15} and improves the local feature representation by disregarding areas of the face that do not contribute to facial muscle movements.

Future work would look into improving the speed of feature extraction and preprocessing. Ideally, a real-time or near real-time system would be able to complete the difference analysis on frames being input from a live video stream. Unfortunately, processing on high-speed video in real-time can be very computationally intensive and not easily accessible for anyone lacking enough funds. It would also be useful to quantify the sensitivity of the system using simulated head movements and find the error rate or methods to counter the effect of head movements. 

Being able to compare with other micro-facial expression datasets that contain baseline sequences is an approach that should be explored. Currently, the SAMM dataset is the only known dataset to provide enough data to produce baseline sequences~\cite{Da16} as other micro-expression datasets, such as SMIC~\cite{Li13a}, CASME I~\cite{Ya13b} and II~\cite{Ya14a}, do not have the raw captured sequences available for similar use. The CASME II baseline sequences in this paper had to be obtained through the excess frames provided after a movement had occurred. To thoroughly compare with similar data contained in SAMM, further data collection of FACS-coded spontaneous micro-movements is required.


%

%

\ifCLASSOPTIONcompsoc
  \section*{Acknowledgments}
\else
  \section*{Acknowledgment}
\fi
The authors would like to thank the Emotional Intelligence Academy for FACS coding the SAMM dataset to allow for ground truth comparison and Brett Hewitt for providing some technical help.

\ifCLASSOPTIONcaptionsoff
  \newpage
\fi



\bibliographystyle{IEEEtran}
\bibliography{completeBibliography}

\begin{thebibliography}{10}
\providecommand{\url}[1]{#1}
\csname url@samestyle\endcsname
\providecommand{\newblock}{\relax}
\providecommand{\bibinfo}[2]{#2}
\providecommand{\BIBentrySTDinterwordspacing}{\spaceskip=0pt\relax}
\providecommand{\BIBentryALTinterwordstretchfactor}{4}
\providecommand{\BIBentryALTinterwordspacing}{\spaceskip=\fontdimen2\font plus
\BIBentryALTinterwordstretchfactor\fontdimen3\font minus
  \fontdimen4\font\relax}
\providecommand{\BIBforeignlanguage}[2]{{%
\expandafter\ifx\csname l@#1\endcsname\relax
\typeout{** WARNING: IEEEtran.bst: No hyphenation pattern has been}%
\typeout{** loaded for the language `#1'. Using the pattern for}%
\typeout{** the default language instead.}%
\else
\language=\csname l@#1\endcsname
\fi
#2}}
\providecommand{\BIBdecl}{\relax}
\BIBdecl

\bibitem{Ru97}
J.~A. Russell and J.~M. Fern{\'a}ndez-Dols, \emph{The psychology of facial
  expression}.\hskip 1em plus 0.5em minus 0.4em\relax Cambridge university
  press, 1997.

\bibitem{Ek92}
P.~Ekman, ``An argument for basic emotions,'' \emph{Cognition and Emotion},
  vol.~6, pp. 169--200, 1992.

\bibitem{Ek04}
{Paul Ekman}, \emph{Emotions Revealed: Understanding Faces and Feelings}.\hskip
  1em plus 0.5em minus 0.4em\relax Phoenix, 2004.

\bibitem{Ek05}
P.~Ekman and E.~L. Rosenberg, \emph{What the Face Reveals: Basic and Applied
  Studies of Spontaneous Expression Using the Facial Action Coding System
  (FACS)}, ser. Series in Affective Science.\hskip 1em plus 0.5em minus
  0.4em\relax Oxford University Press, 2005.

\bibitem{Fr97a}
M.~G. Frank and P.~Ekman, ``The ability to detect deceit generalizes across
  different types of high-stake lies,'' \emph{Journal of Personality and Social
  Psychology}, vol.~72, pp. 1429--1439, 1997.

\bibitem{Ek01}
P.~Ekman, \emph{Telling Lies: Clues to Deceit in the Marketplace, Politics, and
  Marriage}.\hskip 1em plus 0.5em minus 0.4em\relax Norton, 2001.

\bibitem{Po08}
S.~Porter and L.~Ten~Brinke, ``Reading between the lies identifying concealed
  and falsified emotions in universal facial expressions,'' \emph{Psychological
  Science}, vol.~19, no.~5, pp. 508--514, 2008.

\bibitem{Sh12}
X.-B. Shen, Q.~Wu, and X.-L. Fu, ``\BIBforeignlanguage{English}{Effects of the
  duration of expressions on the recognition of microexpressions},''
  \emph{\BIBforeignlanguage{English}{Journal of Zhejiang University SCIENCE
  B}}, vol.~13, no.~3, pp. 221--230, 2012.

\bibitem{Ya13a}
W.-J. Yan, Q.~Wu, J.~Liang, Y.-H. Chen, and X.~Fu,
  ``\BIBforeignlanguage{English}{How fast are the leaked facial expressions:
  The duration of micro-expressions},''
  \emph{\BIBforeignlanguage{English}{Journal of Nonverbal Behavior}}, vol.~37,
  no.~4, pp. 217--230, 2013.

\bibitem{Ek69}
P.~Ekman and W.~V. Friesen, ``Nonverbal leakage and clues to deception,''
  \emph{Psychiatry}, vol.~32, no.~1, pp. 88--106, 1969.

\bibitem{Fr09b}
M.~G. Frank, C.~J. Maccario, and V.~l. Govindaraju, ``Behavior and security,''
  in \emph{Protecting airline passengers in the age of terrorism}.\hskip 1em
  plus 0.5em minus 0.4em\relax Greenwood Pub. Group, 2009.

\bibitem{Oj96}
T.~Ojala, M.~Pietikainen, and D.~Harwood, ``A comparative study of texture
  measures with classification based on featured distributions,'' \emph{Pattern
  Recognition}, vol.~29, no.~1, pp. 51 -- 59, 1996.

\bibitem{Oj02}
T.~Ojala, M.~Pietik{\"a}inen, and T.~M{\"a}enp{\"a}{\"a}, ``Multiresolution
  gray-scale and rotation invariant texture classification with local binary
  patterns,'' \emph{IEEE Transactions on Pattern Analysis and Machine
  Intelligence}, vol.~24, no.~7, pp. 971--987, Jul 2002.

\bibitem{Zh07a}
G.~Zhao and M.~Pietikainen, ``Dynamic texture recognition using local binary
  patterns with an application to facial expressions,'' \emph{Pattern Analysis
  and Machine Intelligence, IEEE Transactions on}, vol.~29, no.~6, pp.
  915--928, 2007.

\bibitem{Da05}
N.~Dalal and B.~Triggs, ``Histograms of oriented gradients for human
  detection,'' in \emph{CVPR}, vol.~1.\hskip 1em plus 0.5em minus 0.4em\relax
  IEEE, 2005, pp. 886--893.

\bibitem{Ch09}
R.~Chaudhry, A.~Ravichandran, G.~Hager, and R.~Vidal, ``Histograms of oriented
  optical flow and binet-cauchy kernels on nonlinear dynamical systems for the
  recognition of human actions,'' in \emph{Computer Vision and Pattern
  Recognition, 2009. CVPR 2009. IEEE Conference on}, June 2009, pp. 1932--1939.

\bibitem{Os09}
M.~O'Sullivan, M.~G. Frank, C.~M. Hurley, and J.~Tiwana, ``Police lie detection
  accuracy: The effect of lie scenario.'' \emph{Law and Human Behavior},
  vol.~33, no.~6, p. 530, 2009.

\bibitem{Fr09a}
M.~Frank, M.~Herbasz, K.~Sinuk, A.~M. Keller, A.~Kurylo, and C.~Nolan, ``I see
  how you feel: Training laypeople and professionals to recognize fleeting
  emotions,'' in \emph{International Communication Association}, 2009.

\bibitem{Ho92}
H.~C. Hopf, W.~Muller-Forell, and N.~J. Hopf, ``Localization of emotional and
  volitional facial paresis,'' \emph{Neurology}, vol.~42, no.~10, pp.
  1918--1918, 1992.

\bibitem{Co09}
J.~F. Cohn, T.~S. Kruez, I.~Matthews, Y.~Yang, M.~H. Nguyen, M.~T. Padilla,
  F.~Zhou, and F.~De~La~Torre, ``Detecting depression from facial actions and
  vocal prosody,'' in \emph{Affective Computing and Intelligent Interaction and
  Workshops, 2009. ACII 2009. 3rd International Conference on}, Sept 2009, pp.
  1--7.

\bibitem{Ka00}
T.~Kanade, J.~F. Cohn, and Y.~L. Tian, ``Comprehensive database for facial
  expression analysis,'' in \emph{Automatic Face and Gesture Recognition, 2000.
  Proceedings. Fourth IEEE International Conference on}, 2000, pp. 46--53.

\bibitem{Lu10}
P.~Lucey, J.~F. Cohn, T.~Kanade, J.~Saragih, Z.~Ambadar, and I.~Matthews, ``The
  extended cohn-kanade dataset (ck+): A complete dataset for action unit and
  emotion-specified expression,'' in \emph{Computer Vision and Pattern
  Recognition Workshops (CVPRW), 2010 IEEE Computer Society Conference on},
  2010, pp. 94--101.

\bibitem{Ya14d}
M.~H. Yap, H.~Ugail, and R.~Zwiggelaar, ``Facial behavioral analysis: A case
  study in deception detection,'' \emph{British Journal of Applied Science \&
  Technology}, vol.~4, no.~10, p. 1485, 2014.

\bibitem{Li13a}
X.~Li, T.~Pfister, X.~Huang, G.~Zhao, and M.~Pietik\"{a}inen, ``A spontaneous
  micro-expression database: Inducement, collection and baseline.'' in
  \emph{10th IEEE International Conference on automatic Face and Gesture
  Recognition}, 2013.

\bibitem{Ya13b}
W.-J. Yan, Q.~Wu, Y.-J. Liu, S.-J. Wang, and X.~Fu, ``Casme database: a dataset
  of spontaneous micro-expressions collected from neutralized faces,'' in
  \emph{IEEE conference on automatic face and gesture recognition}, 2013.

\bibitem{Ya14a}
W.-J. Yan, X.~Li, S.-J. Wang, G.~Zhao, Y.-J. Liu, Y.-H. Chen, and X.~Fu,
  ``Casme ii: An improved spontaneous micro-expression database and the
  baseline evaluation,'' \emph{PLoS ONE}, vol.~9, no.~1, p. e86041, 01 2014.

\bibitem{Da15}
A.~K. Davison, M.~H. Yap, and C.~Lansley, ``Micro-facial movement detection
  using individualised baselines and histogram-based descriptors,'' in
  \emph{Systems, Man, and Cybernetics (SMC), 2015 IEEE International Conference
  on}, Oct 2015, pp. 1864--1869.

\bibitem{Po09}
\BIBentryALTinterwordspacing
S.~Polikovsky, Y.~Kameda, and Y.~Ohta, ``Facial micro-expressions recognition
  using high speed camera and 3d-gradient descriptor,'' in \emph{3rd
  International Conference on Imaging for Crime Detection and Prevention (ICDP
  2009)}, 2009, pp. 16--21. [Online]. Available:
  \url{http://digital-library.theiet.org/content/conferences/10.1049/ic.2009.0244}
\BIBentrySTDinterwordspacing

\bibitem{Sh11}
M.~Shreve, S.~Godavarthy, D.~Goldgof, and S.~Sarkar, ``Macro- and
  micro-expression spotting in long videos using spatio-temporal strain,'' in
  \emph{2011 IEEE International Conference on Automatic Face Gesture
  Recognition and Workshops (FG 2011)}, 2011, pp. 51--56.

\bibitem{Sh14}
M.~Shreve, J.~Brizzi, S.~Fefilatyev, T.~Luguev, D.~Goldgof, and S.~Sarkar,
  ``Automatic expression spotting in videos,'' \emph{Image and Vision
  Computing}, vol.~32, no.~8, pp. 476 -- 486, 2014.

\bibitem{Da14}
A.~K. Davison, M.~H. Yap, N.~Costen, K.~Tan, C.~Lansley, and D.~Leightley,
  ``Micro-facial movements: An investigation on spatio-temporal descriptors,''
  in \emph{ECCVW}, 2014.

\bibitem{Lu14}
Z.~Lu, Z.~Luo, H.~Zheng, J.~Chen, and W.~Li, ``A delaunay-based temporal coding
  model for micro-expression recognition,'' in \emph{Computer Vision-ACCV 2014
  Workshops}.\hskip 1em plus 0.5em minus 0.4em\relax Springer, 2014, pp.
  698--711.

\bibitem{Wa14b}
S.-J. Wang, W.-J. Yan, X.~Li, G.~Zhao, and X.~Fu, ``Micro-expression
  recognition using dynamic textures on tensor independent color space,'' in
  \emph{ICPR}, 2014.

\bibitem{Wa15b}
S.-J. Wang, W.-J. Yan, X.~Li, G.~Zhao, C.-G. Zhou, X.~Fu, M.~Yang, and J.~Tao,
  ``Micro-expression recognition using color spaces,'' \emph{Image Processing,
  IEEE Transactions on}, vol.~PP, no.~99, pp. 1--1, 2015.

\bibitem{Li15a}
Y.-J. Liu, J.-K. Zhang, W.-J. Yan, S.-J. Wang, G.~Zhao, and X.~Fu, ``A main
  directional mean optical flow feature for spontaneous micro-expression
  recognition,'' \emph{Affective Computing, IEEE Transactions on}, vol.~PP,
  no.~99, pp. 1--1, 2015.

\bibitem{Pa15}
D.~Patel, G.~Zhao, and M.~Pietik{\"a}inen, ``Spatiotemporal integration of
  optical flow vectors for micro-expression detection,'' in \emph{Advanced
  Concepts for Intelligent Vision Systems}.\hskip 1em plus 0.5em minus
  0.4em\relax Springer, 2015, pp. 369--380.

\bibitem{Sh09b}
M.~Shreve, S.~Godavarthy, V.~Manohar, D.~Goldgof, and S.~Sarkar, ``Towards
  macro- and micro-expression spotting in video using strain patterns,'' in
  \emph{2009 Workshop on Applications of Computer Vision (WACV)}, dec. 2009,
  pp. 1 --6.

\bibitem{Vi09}
A.~Vinciarelli, A.~Dielmann, S.~Favre, and H.~Salamin, ``Canal9: A database of
  political debates for analysis of social interactions,'' in \emph{Affective
  Computing and Intelligent Interaction and Workshops, 2009. ACII 2009. 3rd
  International Conference on}, Sept 2009, pp. 1--4.

\bibitem{Mo14}
A.~Moilanen, G.~Zhao, and M.~Pietikainen, ``Spotting rapid facial movements
  from videos using appearance-based feature difference analysis,'' in
  \emph{Pattern Recognition (ICPR), 2014 22nd International Conference on}, Aug
  2014, pp. 1722--1727.

\bibitem{Li15b}
X.~Li, X.~Hong, A.~Moilanen, X.~Huang, T.~Pfister, G.~Zhao, and
  M.~Pietik{\"a}inen, ``Reading hidden emotions: Spontaneous micro-expression
  spotting and recognition,'' \emph{arXiv preprint arXiv:1511.00423}, 2015.

\bibitem{Xu16}
F.~Xu, J.~Zhang, and J.~Wang, ``Microexpression identification and
  categorization using a facial dynamics map,'' \emph{IEEE Transactions on
  Affective Computing}, vol.~PP, no.~99, pp. 1--1, 2016.

\bibitem{Ek78a}
P.~Ekman and W.~V. Friesen, \emph{Facial Action Coding System: A Technique for
  the Measurement of Facial Movement}.\hskip 1em plus 0.5em minus 0.4em\relax
  Palo Alto: Consulting Psychologists Press, 1978.

\bibitem{Co04}
T.~F. Cootes, C.~J. Taylor \emph{et~al.}, ``Statistical models of appearance
  for computer vision,'' 2004.

\bibitem{Me13b}
M.~Inc., ``Face++ research toolkit,'' \url{www.faceplusplus.com}, Dec. 2013.

\bibitem{Gu08}
\BIBentryALTinterwordspacing
M.~Guizar-Sicairos, S.~T.~Thurman, and J.~R.~Fienup, ``Efficient subpixel image
  registration algorithms,'' \emph{Opt. Lett.}, vol.~33, no.~2, pp. 156--158,
  Jan 2008. [Online]. Available:
  \url{http://ol.osa.org/abstract.cfm?URI=ol-33-2-156}
\BIBentrySTDinterwordspacing

\bibitem{Da16}
A.~K. Davison, C.~Lansley, N.~Costen, K.~Tan, and M.~H. Yap, ``Samm: A
  spontaneous micro-facial movement dataset,'' \emph{IEEE Transactions on
  Affective Computing}, vol.~PP, no.~99, pp. 1--1, 2016.

\bibitem{Da07}
K.~Dabov, A.~Foi, and K.~Egiazarian, ``Video denoising by sparse 3d
  transform-domain collaborative filtering,'' in \emph{Proc. 15th European
  Signal Processing Conference}, vol.~1, no.~2, 2007, p.~7.

\bibitem{Pf11a}
T.~Pfister, X.~Li, G.~Zhao, and M.~Pietik{\"a}inen, ``Recognising spontaneous
  facial micro-expressions,'' in \emph{International Conference on Computer
  Vision (ICCV)}, nov. 2011, pp. 1449--1456.

\bibitem{Zh07b}
J.~Zhang, M.~Marsza{\l}ek, S.~Lazebnik, and C.~Schmid, ``Local features and
  kernels for classification of texture and object categories: A comprehensive
  study,'' \emph{International journal of computer vision}, vol.~73, no.~2, pp.
  213--238, 2007.

\bibitem{Pe10}
O.~Pele and M.~Werman, ``\BIBforeignlanguage{English}{The quadratic-chi
  histogram distance family},'' in \emph{\BIBforeignlanguage{English}{Computer
  Vision ECCV 2010}}, ser. Lecture Notes in Computer Science, K.~Daniilidis,
  P.~Maragos, and N.~Paragios, Eds.\hskip 1em plus 0.5em minus 0.4em\relax
  Springer Berlin Heidelberg, 2010, vol. 6312, pp. 749--762.

\bibitem{Ah06}
T.~Ahonen, A.~Hadid, and M.~Pietikainen, ``Face description with local binary
  patterns: Application to face recognition,'' \emph{Pattern Analysis and
  Machine Intelligence, IEEE Transactions on}, vol.~28, no.~12, pp. 2037--2041,
  Dec 2006.

\bibitem{Wa81}
\BIBentryALTinterwordspacing
P.~J. Watson and E.~A. Workman, ``The non-concurrent multiple baseline
  across-individuals design: An extension of the traditional multiple baseline
  design,'' \emph{Journal of Behavior Therapy and Experimental Psychiatry},
  vol.~12, no.~3, pp. 257 -- 259, 1981. [Online]. Available:
  \url{http://www.sciencedirect.com/science/article/pii/0005791681900550}
\BIBentrySTDinterwordspacing

\bibitem{Ha15}
T.~O'Haver, ``Signal processing tools,''
  \url{https://terpconnect.umd.edu/~toh/spectrum/SignalProcessingTools.html}.

\end{thebibliography}
\begin{IEEEbiography}[{\includegraphics[width=1in,height=1.25in,clip,keepaspectratio]{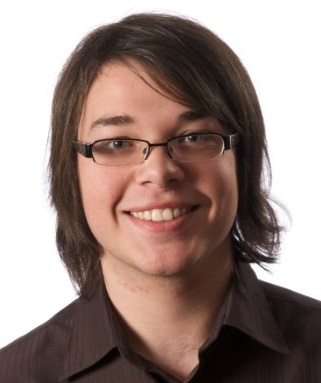}}]{Adrian K. Davison} received his BSc (Hons.) degree in Multimedia Computing in 2012 and PhD in 2016 from the School of Computing, Mathematics and Digital Technology, Manchester Metropolitan University (MMU). He is currently a post-doctoral researcher at MMU within the field of computer vision. He maintains an active role as a representative within MMU. Alongside this role he Co-Chaired the internal MMU Science and Engineering Symposium 2015. His research interests include computer vision, machines learning, facial expression analysis and micro-facial expression analysis.
\end{IEEEbiography}
%
\begin{IEEEbiography}[{\includegraphics[width=1in,height=1.25in,clip,keepaspectratio]{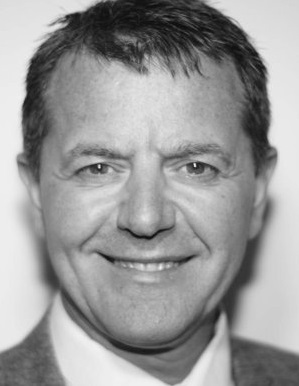}}]{Cliff Lansley}
graduated in education/psychology at Manchester University, UK and has over 25 years experience working at senior levels in public and private sector organisations facilitating leadership, communications, emotional intelligence and coaching programmes. His mission has been to gather the science and push forward the research that can \textquoteleft harden\textquoteright \, the often termed \textquoteleft soft-skill\textquoteright \, of emotional awareness and management (self and others) so that it is embraced more confidently by public/private employers and schools.
\end{IEEEbiography}
\begin{IEEEbiography}[{\includegraphics[width=1in,height=1.25in,clip,keepaspectratio]{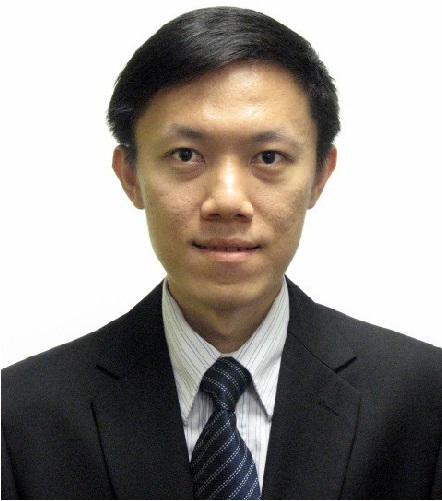}}]{Choon-Ching Ng} received the BSc. (Hons.) and M.Sc. degrees in computer science from Universiti Teknologi Malaysia, in 2007 and 2010, respectively. He received his Ph.D. degree with the School of Computing, Mathematics and Digital Technology, Manchester Metropolitan University in 2015. From 2010 to 2012, he was a Lecturer with University Malaysia Pahang. His research interests include facial aging analysis, pattern recognition, and natural language processing.
\end{IEEEbiography}
\begin{IEEEbiography}[{\includegraphics[width=1in,height=1.25in,clip,keepaspectratio]{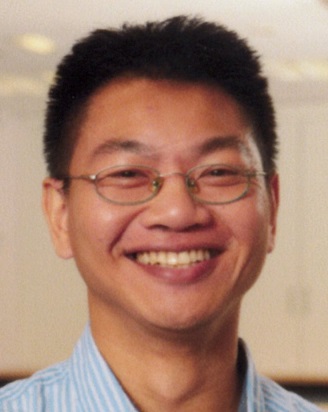}}]{Kevin Tan}
Kevin Tan received his BSc (Hons.) degree in Computer Science (Software Engineering) and MSc.degree in Computer Vision, Visual, and Virtual Environments from University of Leeds, and PhD degree in Bi-manual Interaction within Virtual Environments from Salford University, in 2008. After his PhD, he was a Post-Doctoral Research Assistant with the Materials Science Centre, University of Manchester. He is currently a Senior Lecturer with Manchester Metropolitan University. His research interests are gamification, augmented and virtual reality for cross disciplinary application.
\end{IEEEbiography}
\begin{IEEEbiography}[{\includegraphics[width=1in,height=1.25in,clip,keepaspectratio]{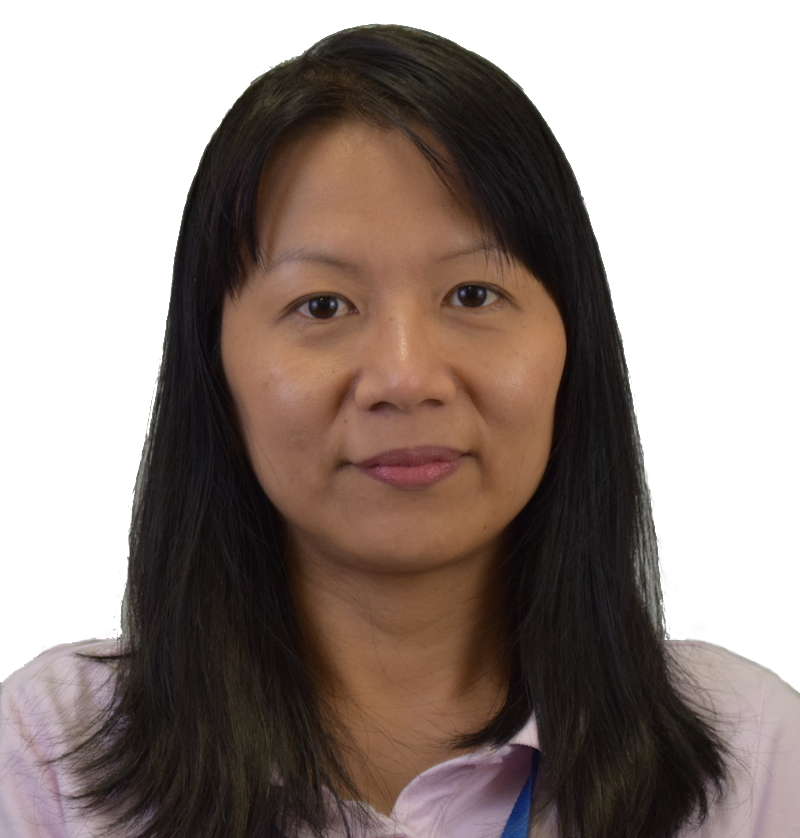}}]{Moi Hoon Yap} is a Senior Lecturer in Computer Science at the Manchester Metropolitan University and a Royal Society Industry Fellow with Image Metrics Ltd.  She received her PhD in Computer Science from Loughborough University in 2009. After her PhD, she worked as Postdoctoral Research Assistant (April 09 - Oct 11) in the Centre for Visual Computing at the University of Bradford. She is a co-Investigator of a Marie Skłodowska-Curie Innovative Training Network, a €3.6M grant awarded by the European Commission to train European experts in multilevel bioimaging, analysis and modelling of vertebrate development and disease (ImageInLife). With the support of two Royal Society International Exchange Schemes, she is collaborating with National University of Science and Technology Taiwan (co-funded by Ministry of Science and Technology Taiwan) and Massey University, New Zealand. She serves as an Associate Editor for Journal of Open Research Software and reviewers for IEEE transactions/journals (Image Processing, Multimedia, Cybernetics, biomedical health and informatics). Her research expertise is in facial analysis and medical image analysis.
\end{IEEEbiography}




\end{document}